\documentclass[10pt,twocolumn,letterpaper]{article}

\usepackage{cvpr}              

\usepackage[table]{xcolor}

\definecolor{cvprblue}{rgb}{0.21,0.49,0.74}
\definecolor{LighterGray}{gray}{0.93}
\definecolor{CustomGreen}{RGB}{153,216,201}
\definecolor{LightGreen}{RGB}{232,244,234}

\usepackage[pagebackref,breaklinks,colorlinks,allcolors=cvprblue]{hyperref}
\usepackage{makecell}
\usepackage{pifont}
\usepackage{multirow}
\usepackage{arydshln}
\usepackage{booktabs}
\usepackage{float}

\usepackage{graphicx}        
\usepackage{caption}         
\usepackage{afterpage}
\usepackage{needspace}
\usepackage{stfloats}
\usepackage{colortbl}
\usepackage{tabularx}

\usepackage{pifont}
\usepackage{epigraph} 
\usepackage[accsupp]{axessibility}

\newif\ifclean

\newcommand{\tablestyle}[2]{\setlength{\tabcolsep}{#1}\renewcommand{\arraystretch}{#2}\centering\footnotesize}
\newcommand{\thickhline}{\Xhline{2\arrayrulewidth}}

\newcommand{\ours}{CamFormer}
\newcommand{\best}[1]{\textbf{#1}} 
\newcommand{\sbest}[1]{\underline{#1}}

\newcommand{\gr}[1]{\textcolor{gray}{#1}}
\newcommand{\gain}[1]{\textcolor{ForestGreen}{+#1}}
\newcommand{\greencheckmark}{\textcolor{ForestGreen}{\ding{51}}}
\newcommand{\CR}[1]{\ifclean{#1}\else {\color{black}{#1}}\fi}
\usepackage{tcolorbox}
\tcbset{colback=gray!10, colframe=gray!10, boxrule=0pt, arc=1pt, left=0pt, right=0pt, top=0pt, bottom=0pt}

\newcommand{\Q}[1]{%
  \par\vspace{2pt}%
  \noindent
  {{\textbf{Q: #1}}}%
  \vspace{2pt}%
}
\newcommand{\para}[1]{
  \par
  \vspace{2pt}
  \noindent\textbf{#1}
}
\newcommand{\custompar}[1]{%
  \par\vspace{3pt}%
  \noindent%
  \colorbox{gray!10}{{#1}}%
  \vspace{2pt}%
}

\definecolor{cvprblue}{rgb}{0.21,0.49,0.74}
\usepackage[pagebackref,breaklinks,colorlinks,allcolors=cvprblue]{hyperref}

\usepackage[flushmargin]{footmisc}

\def\paperID{} 
\def\confName{CVPR}
\def\confYear{2026}

\title{Seeing without Pixels: Perception from Camera Trajectories}

\author{Zihui Xue\textsuperscript{1$^\star$, 2}\quad Kristen Grauman\textsuperscript{2}\quad Dima Damen\textsuperscript{1}\quad Andrew Zisserman\textsuperscript{1}\quad Tengda Han\textsuperscript{1} \\
$^{1}$Google DeepMind\quad$^{2}$The University of Texas at Austin
}

\begin{document}
\maketitle
{
    \renewcommand{\thefootnote}{$\star$}
    \footnotetext{Work done during internship at Google DeepMind.}
}
\begin{abstract}


Can one perceive a video's content without seeing its pixels, just from the camera trajectory—the path it carves through space? This paper is the first to systematically investigate this seemingly implausible question. Towards this end, we propose a contrastive learning framework to train \ours, a dedicated encoder that projects camera pose trajectories into a joint embedding space, aligning them with natural language. We find that, contrary to its apparent simplicity, the camera trajectory is a remarkably informative signal to uncover video content. In other words, ``how you move'' can indeed 
\CR{provide valuable cues about} ``what you are doing'' (egocentric) or ``observing'' (exocentric). We demonstrate the versatility of our learned \ours~embeddings on a diverse suite of downstream tasks, ranging from cross-modal alignment to classification and temporal analysis. Importantly, our representations are robust across diverse camera pose estimation methods, including both high-fidelity multi-sensored and standard RGB-only estimators. Our findings establish camera trajectory as a lightweight, robust, and versatile modality for perceiving video content. 
\footnote{Project webpage: \url{https://sites.google.com/view/seeing-without-pixels}.}
\end{abstract}    
\vspace*{-1mm}
\section{Introduction}
\label{sec:intro}

\begin{figure}[t]
    \centering 
    \includegraphics[width=1.0\linewidth]{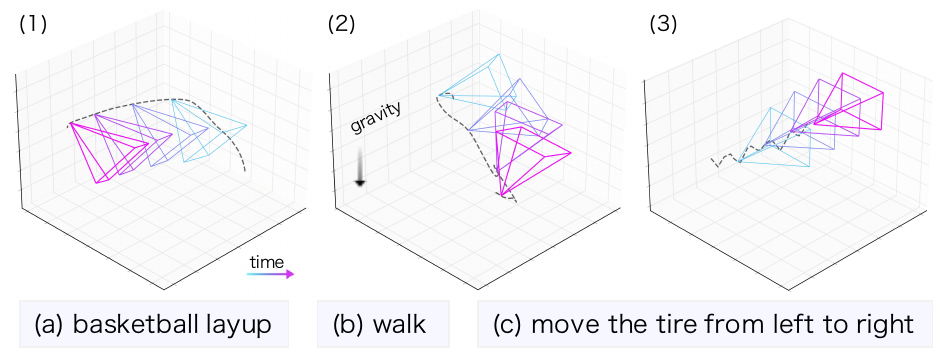}
    \vspace*{-6mm}
    \caption[Can you guess which action goes with which camera pose trajectory?]{
        \textbf{Can you guess which action goes with which camera pose trajectory?}
        In this paper, we find that a camera trajectory carries rich information about the video's content, 
        in both egocentric and exocentric settings.
        Answers are given in the next page.\footnotemark
    }
    \label{fig:teaser}
\end{figure}

\renewcommand{\epigraphflush}{flushleft}
\renewcommand{\epigraphsize}{\footnotesize}
\setlength{\epigraphwidth}{0.45\textwidth}
{\scriptsize
\epigraph{\textit{One sees the environment not with the eyes but with the eyes-in-the-head-on-the-body-resting-on-the-ground.}\\
\hspace{160pt}\textit{James J. Gibson}}
{}
}
\vspace*{-5mm}
We start with a compelling, perhaps even counter-intuitive, question: can a camera's trajectory, devoid of all pixels, be informative enough to reveal the content of a video? Consider the challenge in Fig.~\ref{fig:teaser}. At first glance, matching these simple, abstract curves to specific human actions seems difficult. The key to this puzzle lies in a fundamental principle: human perception is active. We move to see, turning the visual input into an intentional sub-sampling of our surroundings. This principle extends directly to digital capture. Every video is a human-guided sub-sampling of the world, structured by the creator's intent. This intent is what creates the distinct, recognizable motion signatures in Fig.~\ref{fig:teaser} [spoiler alert]: the upward tilt for a basketball layup, the downward left-to-right sweep for moving a tire, and the rhythmic, forward-moving oscillation for walking, are all potential fingerprints of the semantic action, physically written into the trajectory. 

Despite its potential, the intentional signal in camera trajectories has been overlooked. 
The dominant paradigm learns high-level semantics by training visual encoders on massive-scale video-text data via contrastive learning~\cite{wang2022internvideo,wang2024internvideo2,lin2022egocentric,pramanick2023egovlpv2,zhao2023learning,ashutosh2023hiervl}. 
While other non-visual modalities like audio~\cite{guzhov2022audioclip,zhang2025day2dark}, IMU~\cite{moon2023imu2clip,das2025primus,moon2024anymal}, thermal, depth images~\cite{girdhar2023imagebind} and touch~\cite{yang2024binding} have been explored to replace or complement vision, they often require specialized hardware and thus cannot be obtained retroactively from existing videos. In contrast, a camera trajectory, the continuous sequence of camera poses (rotation and translation) over time, is a lightweight 
signal that can be estimated directly from the video itself. We therefore propose to elevate the camera trajectory as a novel modality for video perception, both on its own and as a powerful complement to vision.

The camera trajectory is, of course, not a new concept in computer vision. However, 
it has been commonly used as a \emph{geometric} tool, for tasks such as 3D reconstruction and visual odometry~\cite{hartley2003multiple,li2025megasam,wang2025pi,huang2025vipe,wang2025vggt,li2025egom2p}.
This historical focus has left a fundamental question unexplored: What \emph{semantic} information, if any, is encoded within the camera trajectory itself? 
{To our knowledge, the camera trajectory has received limited attention as a direct source of evidence for perceiving video content,
a gap we address in this paper.}

However, learning to ``interpret'' camera trajectories
presents fundamental challenges. The primary prerequisite—access to high-quality data—has long been a blocker. High-quality poses were historically difficult to obtain: hardware-based solutions were not portable or had low sampling rates~\cite{zhang2012microsoft,keselman2017intel}, while conventional estimation methods were computationally expensive or struggled with accuracy~\cite{agarwal2011rome,davison2007monoslam,mur2015orbslam,schoenberger2016colmap}. Furthermore, the camera trajectory is a low-dimensional, information-sparse signal, posing a conceptual challenge whether it can be informative to disambiguate the many actions a trajectory might represent.

{With regards to the data acquisition challenge, fortunately, recent advances in high-fidelity hardware~\cite{engel2023project} and accurate pose estimation methods~\cite{li2025megasam,wang2025pi,huang2025vipe,wang2025vggt} have made large-scale, high-quality camera trajectory data accessible for the first time, creating the conditions for this modality to emerge. With large-scale paired trajectory-text data now at hand,} we propose to pre-train a dedicated trajectory encoder, which we term {\em \ours}, to project camera trajectories into a joint embedding space with text, using a contrastive learning framework. 
Here, the text refers to action narrations or descriptive video captions detailing the video's content.
Next, to address the semantic ambiguity inherent in camera trajectories, we propose a contextualized trajectory encoding that incorporates extended temporal context for disambiguation.

We structure our investigation to cover two distinct scenarios. We first analyze the \emph{egocentric (first-person)} setting, where a wearable camera's trajectory offers a direct correlation with the recorder's action.
We then analyze 
\emph{exocentric (third-person)} videos, where the trajectory is decoupled from the actor, and reflects the recorder's attention as an observer. To investigate both domains, our evaluation is structured around three core capabilities (cross-modal alignment, classification, and temporal analysis) and spans multiple dimensions, including semantic granularity (coarse activities, fine keysteps) and task types (retrieval, classification, localization). {Across this comprehensive suite of 10 tasks on 5 datasets, \ours~effectively unlocks the semantic potential of the camera trajectory, delivering consistent gains ranging from +3.2\% to +13.2\%. The trajectory proves to be a powerful standalone signal: our lightweight \ours~is capable of outperforming computationally-heavy vision models in key scenarios. It also excels as a valuable complementary signal, providing the best overall performance when fused with vision.} Finally, we show that \ours~is robust across various camera pose sources, from high-fidelity multi-sensor SLAM to standard video-only estimates, demonstrating its practical utility.

\section{Related Work}
\label{sec:related_work}
\para{Multimodal Contrastive Learning.} Our world is inherently multimodal.
The success of vision-language models like CLIP~\cite{radford2021learning} establish a de facto standard for binding our rich visual world to the semantic structure of text~\cite{lin2022egocentric,pramanick2023egovlpv2,wang2022internvideo,wang2024internvideo2}, using a contrastive objective~\cite{oord2018representation} that aligns corresponding pairs in a shared embedding space. 
The same principle has been applied to a broader set of modalities, seeking to connect audio~\cite{guzhov2022audioclip}, IMU~\cite{moon2023imu2clip,das2025primus,moon2024anymal}, thermal, depth images~\cite{girdhar2023imagebind} and touch~\cite{yang2024binding} to a shared semantic space. Conspicuously absent from this list, however, is the camera pose trajectory, despite being an intrinsic property of any video recording. Our work addresses this gap, introducing it as a new modality and demonstrating its immense potential for semantic representation learning through a series of downstream tasks.

\footnotetext{{\rotatebox{180}{Answer: (1)-(c), (2)-(a), (3)-(b)}}}
\para{Action Understanding with Egocentric Motion.} Motion has long been recognized as a helpful signal in egocentric videos. A few works~\cite{ryoo2015pooled,singh2017trajectory,abebe2016robust,narayan2014action} have shown that egocentric motion representations, such as optical flow~\cite{kitani2011fast,kitani2012ego,wang2013action,li2015delving,singh2015generic}, can aid action recognition. Another line of work~\cite{stikic2008adl,lyons2021improved,padmanabha2025egocharm,ehsani2020can,tan2023egodistill,zhang2024masked,yang2025reading} captures motion signals explicitly using IMU sensors mounted on the camera wearer's head or limbs. Beyond action recognition, egocentric motion has also been shown to correlate with other semantic properties, such as physical forces~\cite{park2016force} and camera-wearer identity~\cite{yonetani2017ego}. While these works support the premise that egocentric motion is informative, they are confined to the egocentric domain and a narrow set of tasks. Our investigation takes a broader view by systematically investigating the camera trajectory, across both egocentric and exocentric domains, and on a diverse suite of downstream tasks.

\para{Camera Pose in 3D Vision.} Camera pose estimation is a fundamental task of 3D vision. Recent advances deliver increasingly precise trajectories, driven by both multi-sensor hardware like Meta Aria glasses~\cite{engel2023project} and learning-based models~\cite{li2025megasam,huang2025vipe,wang2025pi,wang2025vggt,xiao2025spatialtrackerv2,kuang2024collaborative} that infer pose from monocular videos. Our work is orthogonal to these efforts: we focus on interpreting the resulting trajectories, with performance naturally benefiting as estimation continues to improve. On the application side, camera pose has powered tasks such as novel view synthesis~\cite{li2025cameras}, 3D reconstruction and mapping~\cite{keetha2025mapanything}, or serving as a conditional prior to guide video generation~\cite{he2024cameractrl,li2024can,li2025egom2p,geng2025motion}. 
{While works on 3D human motion estimation~\cite{jiang2017seeing,yuan20183d,luo2021dynamics,li2023ego,yi2025estimating} and hand forecasting~\cite{Hatano2025EgoH4} share a similar premise—that camera motion is a signature of the actor's movement—their goal remains the estimation of the physical body / hands. Our work is the first to broaden this perspective, proposing that the trajectory encodes rich semantic signals for video perception.}

\para{Generating Camera Motion Descriptions.} Recent work~\cite{fang2025camerabench,liu2025shotbench,wang2025cinetechbench} trains large multimodal models (LMMs) to generate textual descriptions of camera cinematography from video input (\eg, ``zoom, ``pan''). While these works treat camera motion as a video attribute to describe, we treat the trajectory as a semantic signal to interpret, which our experiments prove better decodes video content.

\section{Method}
\label{sec:method}

\begin{figure*}[t]
\centering
\includegraphics[width=1.0\textwidth]{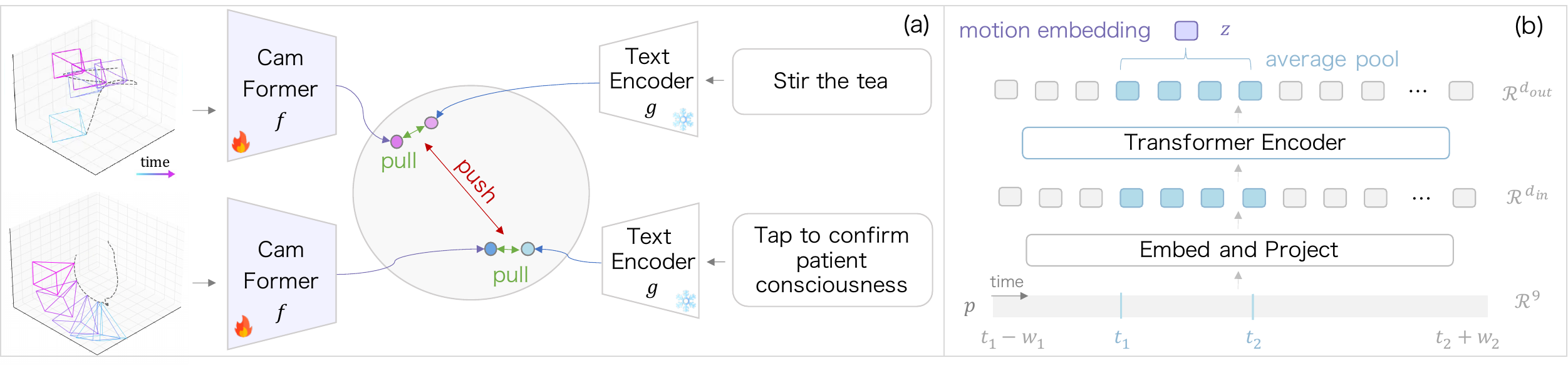}
\vspace{-5mm}
\caption{\small\textbf{Unlocking Semantic Information Hidden in Camera Trajectories.} (a) We propose contrastive pre-training on paired (trajectory, text) data. Our model, \ours, is trained to map camera trajectories into a joint semantic space, aligning them with natural language. (b) We propose contextualized trajectory encoding that incorporates extended temporal context 
to disambiguate the local action. 
}
\vspace{-3mm}
\label{fig:method}
\end{figure*}


This work addresses the untapped problem of how to unlock the semantic information within camera pose trajectories. Our core methodology is to learn a dedicated encoder to map camera trajectories into a shared semantic space with text (Sec.~\ref{sec:method_pretraining}). We then apply the learned camera trajectory embeddings to a suite of downstream tasks to assess their utility for perceiving video content (Sec.~\ref{sec:method_application}).


\subsection{Learning Camera Trajectory Embeddings} 
\label{sec:method_pretraining}
We apply 
multimodal contrastive learning~\cite{radford2021learning} to a new modality pairing: the camera trajectory of a video clip and the text description of that clip's content. 

\para{Problem Formulation.} Formally, let $\mathbf{v}$ be a video clip spanning the time interval $[t_1, t_2]$, and $\mathbf{t}$ be its paired text description. Let $\mathbf{p} \in \mathbb{R}^{N\times9}$ denote the corresponding camera pose trajectory derived from $\mathbf{v}$, where $N$ is the number of pose samples extracted from the clip's duration $t_2 - t_1$ at a given sampling rate $s$,~\ie, $N = (t_2 - t_1) \times s$. Each pose is represented as a 9D relative vector (3D translation and 6D continuous rotation representation~\cite{zhou2019continuity}), computed with respect to the sequence midpoint. The representation choice is justified in our ablation (cf. Supplementary). 



\para{Training Objective.} We employ InfoNCE~\cite{oord2018representation} loss, where the matching pairs of camera trajectory and text in a batch are treated as positives and all other pairwise combinations in the batch are regarded as negatives. For a batch of $B$ examples, $\{(\mathbf{p}_i, \mathbf{t}_i)\}_{i=1}^B$, we optimize the loss $\mathcal{L} = \mathcal{L}_{P \to T} + \mathcal{L}_{T \to P}$:
\begin{equation*}
\vspace{-2mm}
\mathcal{L}_{P \to T} = 
-\frac{1}{B} \sum_{i=1}^B 
\log 
\frac{
    \exp\!\left( 
        f(\mathbf{p}_i) \cdot g(\mathbf{t}_i) / \tau 
    \right)
}{
    \sum_{j=1}^B 
    \exp\!\left( 
        f(\mathbf{p}_i) \cdot g(\mathbf{t}_j) / \tau 
    \right)
}
\end{equation*}
where $\mathcal{L}_{T \to P}$ is the symmetric text-to-trajectory loss, and $\tau$ is the temperature hyperparameter. The text encoder $g$ is the pre-trained and frozen encoder from CLIP~\cite{radford2021learning}, which provides a fixed semantic target and allows $f$ to learn representations grounded in its robust embedding space.

\para{Model Architecture.} Our camera trajectory encoder $f$ (\ours), 
is a light-weight Transformer~\cite{vaswani2017attention}. The input 9D pose sequence $\mathbf{p}\in\mathbb{R}^{N\times 9}$ is first linearly projected into $d_{\text{in}}$-dimensional embeddings $\mathbb{R}^{N\times d_{\text{in}}}$. After adding positional embeddings, the full sequence is processed by a series of Transformer blocks to fuse temporal information. Finally, the output features are temporally mean-pooled to a vector in $\mathbb{R}^{d_{\text{in}}}$, and projected by a linear layer to $d_{\text{out}}$ dimensions to produce the final vector $z = f(\mathbf{p}) \in \mathbb{R}^{d_{\text{out}}}$. 

\para{Contextualized Trajectory Encoding.} A unique challenge in encoding camera trajectories is the low information density.
Comparing with a short video clip (\eg 1 second),
a camera trajectory with a similar duration carries much sparser information and could be semantically ambiguous.
We address this challenge with a contextualized trajectory encoding,  which extends the sequence length beyond the immediate temporal window to capture broader temporal context, thereby disambiguating the central action.

Formally, we extend the base window $[t_1, t_2]$ by a total duration $w$. $w$ is then randomly split into $w_1$ and $w_2$ (such that $w_1 + w_2 = w$), resulting in an extended, temporally-shifted window $[t_1 - w_1, t_2 + w_2]$. The trajectory in this extended window serves as the final input to our \ours~$f$. After $f$ processes this entire sequence, the final embedding $\mathbf{z}$ is produced by mean-pooling only the $N$ output features corresponding to the original window $[t_1, t_2]$. 
This strategy infuses the local representation with global context without diluting it with potentially irrelevant adjacent actions. 
The encoding process is shown in Fig.~\ref{fig:method} (b).

\subsection{Applications\,of\,Camera\,Trajectory\,Embeddings} \label{sec:method_application}
After pre-training \ours~$f$ on paired camera trajectories and text,
we test the power of its representation across a diverse suite of downstream tasks: 
\begin{itemize}
    \item \textbf{Cross-modal Alignment.} Through text retrieval, we directly test the learned alignment between a camera trajectory and the video's content, as represented by its paired text (\ie, action narrations or video captions). 
    \item \textbf{Downstream Classification.} We assess our embeddings on a spectrum of classification tasks. {This includes identifying coarse-grained scenarios reflecting the overall activity (\eg, distinguishing between cooking, dancing, or basketball) and fine-grained keysteps (\eg, within a cooking activity, identifying cutting, pouring water and washing fruit).} 
    Additionally, we evaluate the ability to discern motion signatures indicative of performer skill levels via proficiency estimation (\ie, beginner or expert).
    \item \textbf{Temporal Analysis.} We examine tasks that require precise temporal information, including temporal action localization and the recognition of periodic patterns for repetitive action counting.
\end{itemize}

\noindent The pre-trained \ours~is applied to these diverse tasks using two standard evaluation paradigms: (1) as a frozen feature extractor (\eg, linear probing); 
and (2) via end-to-end fine-tuning, where \ours~is trained jointly with a linear head. See Supp. for details.

\section{Experimental Setup}\label{sec:exp_setup}

\begin{table}[!t]
\tablestyle{4pt}{1.2}
\caption{\small \textbf{Datasets Used in Our Analysis}. Black checkmarks (\ding{51}) denote data provided by the original dataset, while green checkmarks (\greencheckmark) reflect our enhancement (estimated camera trajectories). The subscript $\times n$ specifies the number of available trajectories sources. We provide three additional pose sources (MegaSaM~\cite{li2025megasam}, ViPE~\cite{huang2025vipe} and $\pi^3$~\cite{wang2025pi}) to enrich Ego-Exo4D and UCF101, and one ($\pi^3$~) to FineGym. ``label'' denotes downstream task annotations (\eg, action labels). The \colorbox{LightGreen}{green-highlighted rows} denote our pre-training datasets.}
\vspace{-3mm}
\begin{tabular}{lcccc}
\thickhline
Dataset & Domain & Trajectory & {Text} & \# Hrs \\
\hline
\rowcolor{LightGreen}
Ego-Exo4D~\cite{grauman2024ego} & ego & \ding{51}\greencheckmark\textsubscript{\textcolor{ForestGreen}{$\times$3}} & narration \& label & 221.3 \\ 
Nymeria~\cite{ma2024nymeria} & ego & \ding{51} & narration & 38.6 \\
\rowcolor{LightGreen} 
DynPose-100K~\cite{rockwell2025dynamic} & exo & \ding{51}\textsubscript{$\times$2} & caption & 157.5 \\
UCF101~\cite{soomro2012ucf101} & exo & \greencheckmark\textsubscript{\textcolor{ForestGreen}{$\times$3}} & label & 27.0 \\
FineGym~\cite{shao2020finegym} & exo & \greencheckmark\textsubscript{\textcolor{ForestGreen}{$\times$1}} & label & 92.8 \\
\thickhline
\end{tabular}
\label{tab:dataset}
\end{table}
\vspace*{3mm}

\begin{figure*}
    \centering
    \includegraphics[width=1.0\textwidth]{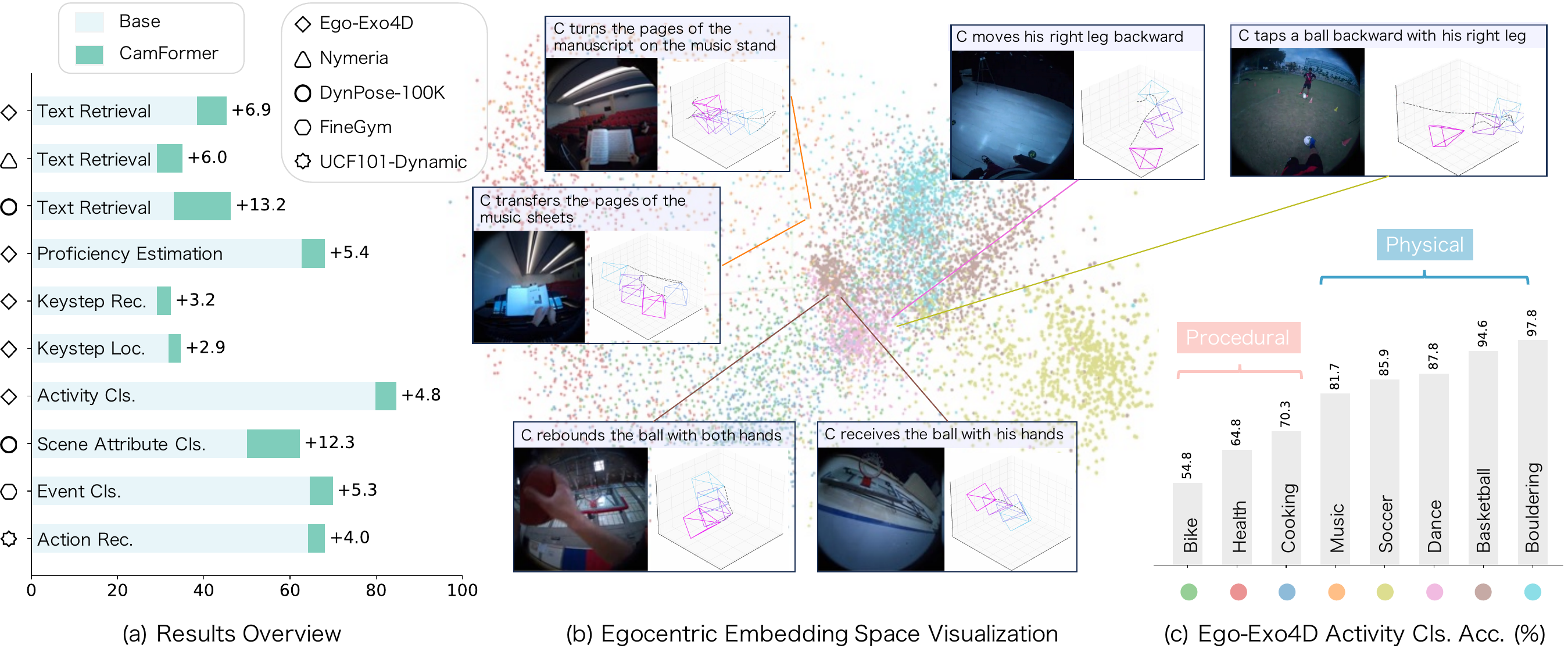}
    \vspace*{-8mm}
    \caption{\small 
    {(a) \textbf{Quantitative Results Overview}: we summarize \ours's performance against base methods / models on 10 downstream tasks across 5 datasets, demonstrating its consistent performance advantages; (b) \textbf{A PCA visualization of \ours~embeddings} on unseen Ego-Exo4D trajectories, colored by the dataset's 8 activity labels. (Note: \ours~only takes the trajectory as input; video clips and text are shown for interpretation only); (c) Per-class activity classification accuracy plot on Ego-Exo4D reveals a performance dichotomy: \ours~excels at physical activities but is less effective on procedural ones with more subtle camera motions.}
    }
    \label{fig:results_overview}
\end{figure*}

\para{Bridging the Data Gap.} 
We observe a trajectory-semantic data gap: traditional 2D video understanding datasets are vast and semantically rich but lack camera pose trajectories~\cite{soomro2012ucf101,grauman2022ego4d,damen2018scaling,shao2020finegym}, while 3D datasets that include poses often focus on static scenes and navigation tasks~\cite{dai2017scannet,chang2017matterport3d,habitat19iccv}, rather than depicting diverse human actions or high-level semantics. Fortunately, recent advances in both hardware capture and software-based pose estimation~\cite{engel2023project,li2025megasam,wang2025pi,wang2025vggt,huang2025vipe} allow us to source and assemble these components. Table~\ref{tab:dataset} summarizes our data effort to bridge the gap.

As shown in the table, we adopt two large-scale datasets for pre-training. For the egocentric domain, we adopt the large-scale Ego-Exo4D~\cite{grauman2024ego}, which provides time-stamped narrations and high-quality dense camera trajectories from Meta Aria glasses~\cite{engel2023project} (specifically, visual-inertial pose estimates from the device's SLAM cameras). As these are head-mounted, the camera trajectory serves as a direct and high-fidelity proxy for the wearer's head motion. We only use its egocentric videos, as the dataset's exocentric cameras are static and thus uninformative for our motion-based analysis. {For the exocentric domain, we pre-train on DynPose-100K~\cite{rockwell2025dynamic} data, where we utilize video captions from Panda70M~\cite{chen2024panda} and the two available pose sources (the original one provided by the dataset and an alternate one estimated by ViPE~\cite{huang2025vipe}).}
Our downstream evaluation suite includes both Ego-Exo4D and DynPose-100K, supplemented by Nymeria~\cite{ma2024nymeria} {(another egocentric dataset collected by Aria glasses for full-body motion understanding)}
and two standard exocentric action recognition datasets (UCF101~\cite{soomro2012ucf101} and FineGym~\cite{shao2020finegym}). 

We enrich these datasets with estimated camera trajectories. On Ego-Exo4D, we supplement the dataset's original camera poses with trajectories from three leading pose estimators (MegaSaM~\cite{li2025megasam}, ViPE~\cite{huang2025vipe}, and $\pi^3$~\cite{wang2025pi}). This multiple-source testbed not only allows us to benchmark model robustness across pose sources, but also offers a novel, semantic-based evaluation for the pose estimators themselves (\ie, which estimator yields the best downstream semantic performance, cf. Table~\ref{tab:egoexo4d_camest_combined}). We also apply this estimation pipeline to UCF101 and FineGym to generate the pose trajectories that they lack, thereby enabling our analysis on these standard action benchmarks.

\para{Implementation.} 
We sample input camera poses at 5-30 Hz, subject to datasets.
For our \ours, we set the internal projection dimension $d_{in}$ to 128 and the final output dimension $d_{out}$ to 512, which matches our frozen text CLIP encoder. The transformer encoder consists of 4 layers, each with 4 attention heads, a 256-dimensional feed-forward network, and a dropout rate of 0.1.
During training, the context duration $w$ is uniformly sampled from $\mathcal{U}(0, w_{\text{max}})$, where $w_{\text{max}}=8s$ is the maximum context length.
During inference, we test various values of $w$ to analyze the impact of temporal context (cf. Sec.~\ref{sec:exp_results_analysis}). 

\section{Results} \label{sec:exp_results}
We structure our findings as a series of questions, exploring the egocentric (Sec.~\ref{sec:exp_results_ego}) and exocentric (Sec.~\ref{sec:exp_results_exo}) domains, followed by further analysis (Sec.~\ref{sec:exp_results_analysis}). Fig.~\ref{fig:results_overview} (a) provides a high-level results overview,
comparing \ours~against base methods / models across 10 downstream tasks \CR{encompassing diverse activities types}, and highlighting its consistent gains that we will detail throughout this section.

\subsection{\ours\ for Egocentric Videos}\label{sec:exp_results_ego}
{Our analysis in the egocentric domain reveals a strong, direct link between the camera's trajectory and the recorder's activity, which our \ours~effectively captures. We see this in Fig.~\ref{fig:results_overview} (b), which visualizes PCA embeddings of \ours's features on the unseen Ego-Exo4D validation set. The embeddings demonstrate some natural clustering by the dataset's activity labels (a task for which our model was not explicitly trained). Furthermore, the visualized data pairs show that trajectories close in the embedding space share action semantics, even if their specific motion patterns or overall activity labels are different. This demonstrates that \ours~learns to decipher the underlying semantic meaning encoded within a camera trajectory through our pre-training.}

\Q{When is a camera trajectory most informative?}  

{\noindent To quantify the emergent clustering observed in the embedding space, we fine-tune \ours~for the downstream activity classification task on Ego-Exo4D using its 8 activity labels. The per-class accuracy shown in Fig.~\ref{fig:results_overview} (c) reveals a clear pattern: our model excels at recognizing dynamic physical activities, achieving accuracies over 90\% for basketball and bouldering, but gets confused among the three procedural activities (i.e., bike repair, health and cooking), where camera motions are more localized and subtle. This observation forms the basis of our analysis below.}

\begin{figure}[!t]
  \centering
   \includegraphics[width=1.0\linewidth]{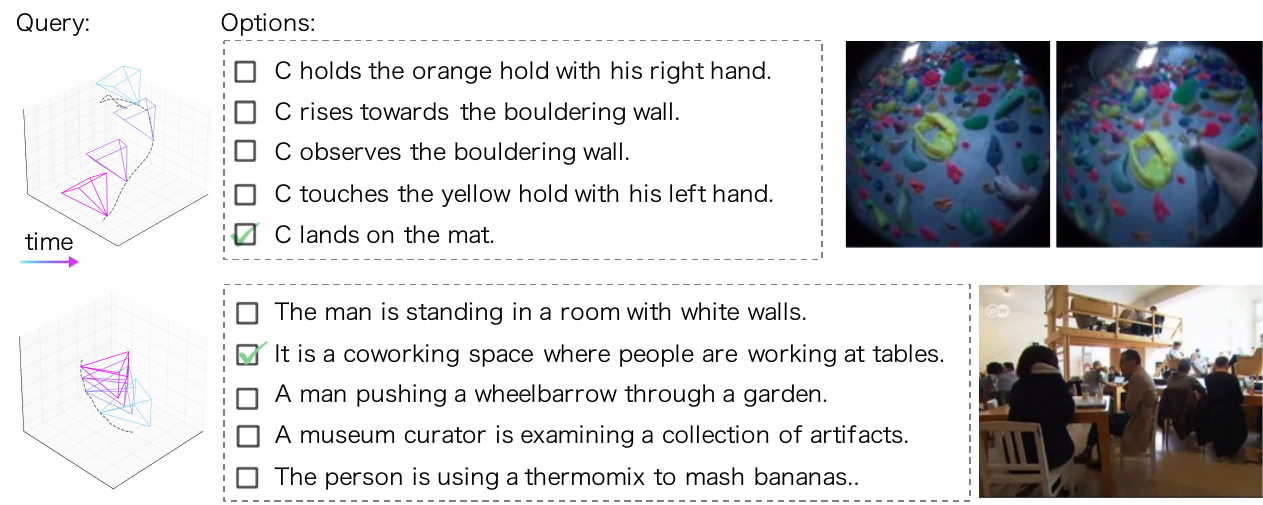}
   \caption{\small \textbf{Qualitative Text Retrieval Results} on egocentric Ego-Exo4D (up) and exocentric DynPose-100K (bottom). 
   Up: A clear downward pose trajectory disambiguates the action of landing, where visual cues are subtle. 
   Bottom: A circling trajectory, common for capturing a scene overview, is correctly associated with the high-level scene description. See Supp. for more qualitatives.
   }
   \label{fig:qualitative_retrieval}
\end{figure}

\Q{How do camera trajectories compare to other modalities?}
\custompar{Evaluation Setup} 
{To serve as our main evaluation for cross-modal alignment, we create a new, challenging text retrieval benchmark from the Ego-Exo4D validation split, comprising 7079 camera trajectory queries. To reduce retrieval ambiguity, we adapt the 5-way multiple-choice question (MCQ) format from~\cite{lin2022egocentric}, where the model must select the correct textual description from five candidates. The queries are carefully balanced across the dataset's 8 activity labels and its ``in-view'' (iv) / ``out-of-view'' (oov) visibility annotations to facilitate more detailed analysis. Fig.~\ref{fig:qualitative_retrieval} (up) provides an example of this task, showing an ``oov'' case where the action is not directly visible in the egocentric video (see Supp. for more).}


Our primary comparison is with established contrastive multimodal methods incorporating common modalities aligned with text: 
CLAP (audio-text)~\cite{elizalde2024natural}, PRIMUS (IMU-text)~\cite{das2025primus}, CLIP (image-text)~\cite{radford2021learning} and EgoVLPv2 (video-text)~\cite{pramanick2023egovlpv2}. We report on two EgoVLPv2 checkpoints (pre-trained on Ego4D~\cite{grauman2022ego4d} and Ego-Exo4D, respectively) and note that PRIMUS is also pre-trained on Ego-Exo4D, aligning with our setup. As text retrieval is formulated as MCQ, we also report performance of one representative LMM, Gemini-2.5-Pro~\cite{comanici2025gemini}, to highlight the challenging nature of our MCQ task. We stress that this is not a direct comparison due to Gemini's prohibitive cost and different paradigm. 

\custompar{Results} Table~\ref{tab:main_retrieval_results} benchmarks~\ours's performance on our 5-way MCQ task. 
{Remarkably, on physical activities, it outperforms computationally-heavy video baselines with a much lower computational cost, proving its strength as a lightweight standalone signal. On procedural activities, \ours~embeddings provide the best results overall when fused with video (achieved by averaging our camera trajectory features with the best-performing video features, EgoVLPv2~\cite{pramanick2023egovlpv2}).  This confirms the camera trajectory is a versatile signal, effective both on its own and as a complementary source of information for video.}
Fig.~\ref{fig:qualitative_retrieval} (up) provides an ``oov'' example that illustrates its advantage. In the bouldering activity, distinguishing between fine-grained actions like ``rise towards wall'' and ``land on the mat'' is difficult from visual frames alone. However, the camera pose trajectory provides an unambiguous downward signal, allowing our model to correctly identify ``land on the mat.'' This confirms camera trajectory's unique value in disambiguating actions when visual cues are subtle or misleading. 

\Q{How generalizable is the learned \ours?} 
\custompar{Evaluation Setup} 
After \ours~is pre-trained exclusively on Ego-Exo4D~\cite{grauman2024ego},
we test this model by directly applying it on Nymeria~\cite{ma2024nymeria} in a zero-shot manner.
We use the entire Nymeria dataset as a test set and create 4000 questions (1000 for each of its four narration types: legs/feet, focus attention, body posture, hands/arms), following the same MCQ design used for Ego-Exo4D.

\custompar{Results} Table~\ref{tab:main_retrieval_results} right columns show that our model demonstrates strong zero-shot generalization to Nymeria, performing effectively across its diverse annotation types. Relative to baselines, it is particularly strong on narrations describing ``legs'' and ``focus attention''—the categories often being out-of-view, where baseline visual models perform near random guess. This confirms our model's unique ability to understand non-visible actions from their motion signature, echoing our ``oov'' analysis on Ego-Exo4D. 

\begin{table*}[!t]
\tablestyle{5pt}{1.2}
\caption{\small \textbf{Egocentric Text Retrieval Results}. We report top-1 accuracy (\%) for a 5-way MCQ on Ego-Exo4D and Nymeria (zero-shot). Chance is 20\%. Best results are bolded and second best results are underlined. 
For Ego-Exo4D, columns show performance on splits where the narration is in-view (iv) or out-of-view (oov) from the \emph{egocentric} perspective, across both physical and procedural activities. For Nymeria, columns show performance on four motion annotation types (legs/feet, focus attention, body posture, hands/arms). The grayed-out Gemini-2.5-Pro row is included to highlight the challenging nature of our benchmark, not as a direct, apples-to-apples comparison. Key takeaways are: (1) As a standalone signal, the camera trajectory is a strong, low-cost modality that outperforms video on physical activities and in visually challenging scenarios (\eg, ``oov'' and ``legs'' split); (2) As a complementary signal, for cases where vision is strong (\eg, procedural activities), fusing it with video features ($^{\star}$) improves over the video-only baseline, proving its non-redundant value.}
\vspace{-2mm}
\begin{tabular}{lccc|ccccc|ccccc}
\thickhline
\multirow{3}{*}{Method} & \multirow{3}{*}{Modality} & \multirow{3}{*}{\makecell[c]{\# MACs\\(G)}} & \multirow{3}{*}{\makecell[c]{\# Params\\(M)}} & \multicolumn{5}{c|}{Ego-Exo4D~\cite{grauman2024ego}} & \multicolumn{5}{c}{Nymeria~\cite{ma2024nymeria}} \\
& & & & \multicolumn{2}{c}{Physical} & \multicolumn{2}{c}{Procedural} & \multirow{2}{*}{all} & \multirow{2}{*}{legs} & \multirow{2}{*}{focus} & \multirow{2}{*}{body} & \multirow{2}{*}{hands} & \multirow{2}{*}{all} \\
& & & & iv & oov & iv & oov & & & & & & \\
\hline
CLAP~\cite{elizalde2024natural} & audio & 6.80 & 32.8 & 21.4 & 20.1 & 29.5 & 33.3 & 24.6 & - & - & - & - & - \\
PRIMUS~\cite{das2025primus} & IMU & 0.03 & 1.4 & 18.5 & 25.3 & 24.7 & 22.0 & 23.2 & - & - & - & - & - \\
CLIP~\cite{radford2021learning} & image & 2.95 & 59.0 & 25.2 & 18.2 & 26.8 & 21.9 & 22.9 & 22.4 & 22.4 & 40.5 & 29.5 & 28.7 \\
EgoVLPv2~\cite{pramanick2023egovlpv2} (Ego4D) & video & 89.49 & 150.7 & 30.9 & 24.8 & 48.7 & 40.9 & 34.4 & 22.1 & 17.4 & \sbest{45.7} & \sbest{31.7} & 29.2 \\
EgoVLPv2~\cite{pramanick2023egovlpv2} (Ego-Exo4D) & video & 89.49 & 150.7 & 39.1 & 25.6 & \sbest{50.5} & \sbest{45.4} & 38.4 & 22.8 & 20.5 & 39.6 & 28.5 & 27.9 \\
\gr{Gemini-2.5-Pro~\cite{comanici2025gemini}} & \gr{video} & - & - & \gr{53.9} & \gr{29.0} & \gr{67.4} & \gr{50.9} & \gr{48.7} & \gr{32.6} & \gr{36.6} & \gr{60.7} & \gr{52.7} & \gr{45.6} \\
\rowcolor{blue!5}
\ours~embeddings & trajectory & 0.02 & 0.3 & \best{56.1} & \best{46.4} & 34.3 & 32.7 & \sbest{44.8} & \best{30.8} & \best{33.2} & 36.3 & 26.0 & \sbest{31.6} \\
\rowcolor{blue!15}
\ours~embeddings$^{\star}$ & video+trajectory & 89.51 & 151.0 & \sbest{56.0} & \sbest{45.8} & \best{51.4} & \best{45.9} & \best{46.0} & \sbest{30.1} & \sbest{30.6} & \best{47.8} & \best{34.0} & \best{35.6} \\
\thickhline
\end{tabular}
\label{tab:main_retrieval_results}
\end{table*}

\begin{table}[!t]
\tablestyle{2pt}{1.2}
\caption{\textbf{Analysis of Model Robustness}. Our \ours, despite being pre-trained only on Aria poses, generalizes effectively to process various estimated poses. Left (Ego-Exo4D activity classification): Models initialized from our \ours~consistently outperform their counterparts trained from scratch. Right (Ego-Exo4D keystep recognition): Fusing camera trajectory features with video consistently outperforms the video-only baseline (29.17\%), with performance gains shown in brackets. This demonstrates the complementary value of camera trajectory.}
\begin{tabular}{lcccc}
\thickhline
\multirow{2}{*}{Pose Source} & \multicolumn{2}{c}{Activity Cls.} & \multicolumn{2}{c}{Keystep Rec.} \\
 & Pretrain \ding{55} & Pretrain \ding{51} & Trajectory & Video+Trajectory \\
\hline
MegaSaM~\cite{li2025megasam} & 53.67 & 60.83 (\gain{7.16}) & 11.49 & 32.60 (\gain{3.43}) \\
ViPE~\cite{huang2025vipe}    & 60.83 & 66.15 (\gain{5.32}) & 12.26 & 32.21 (\gain{3.04})\\
$\pi^3$~\cite{wang2025pi}     & 61.47 & 66.15 (\gain{4.68}) & 12.76 & 32.83 (\gain{3.66}) \\
\hline
Aria~\cite{engel2023project}                        & 61.83 & 71.28 (\gain{9.45}) & 14.07 & 32.37 (\gain{3.20}) \\
\thickhline
\end{tabular}
\label{tab:egoexo4d_camest_combined}
\end{table}

\Q{{How do camera trajectories perform on physical vs. procedural tasks?}}
\custompar{Evaluation Setup} Based on our activity classification findings (Fig.~\ref{fig:results_overview} (c)), we perform a deeper analysis on the following targeted Ego-Exo4D tasks: (1) proficiency estimation on binary labels of expert vs.~non-expert for the two physical activities (bouldering, dancing); (2) keystep recognition on the 278 fine-grained keystep labels, which are specific to the 3 procedural activities (bike repair, health, cooking); and (3) temporal action localization, using keystep boundaries from the same procedural activities.


\custompar{Results} 
This analysis confirms the pattern from Fig.~\ref{fig:results_overview} (c), revealing a  two-fold role for the camera trajectory:
\begin{itemize}
    \item For \textit{physical activities}, the camera trajectory is a powerful standalone signal. On the proficiency estimation task, \ours~achieves an average +5.4\% gain over the strong video baseline~\cite{bertasius2021space} (Fig.~\ref{fig:results_overview} (a)), effectively capturing the motion signatures of expertise.
    \item For \textit{procedural activities}, where motion patterns can be ambiguous, the camera trajectory provides valuable complementary information. Fusing \ours~features with the strong EgoVLPv2~\cite{pramanick2023egovlpv2} video features on keystep tasks (Fig.~\ref{fig:results_overview} (a)) yields a +3.2\% accuracy gain in recognition and a +2.9 mIoU@0.3 gain in localization. 
\end{itemize}
Full results for these tasks are available in Supplementary.

\begin{table}[!t]
\tablestyle{3pt}{1.2}
\caption{\small \textbf{Exocentric Text Retrieval Results} on DynPose-100K. We report top-1 accuracy (\%) for a 5-way MCQ. Our \ours~not only performs well above random chance (20\%) across both pose sources (original~\cite{rockwell2025dynamic} and ViPE~\cite{huang2025vipe}), confirming a meaningful trajectory-semantic link in the exocentric domain, but also surpasses 
\CR{strong LMM} baselines, demonstrating the effectiveness of interpreting the camera trajectory directly.}
\vspace*{-2mm}
\begin{tabular}{lcc}
\thickhline
Method & Modality & Acc. (\%) \\
\hline
Qwen2.5-VL-7B on CameraBench~\cite{fang2025camerabench} & cam desc. (text) & 27.8 \\
ShotVL-7B~\cite{liu2025shotbench} & cam desc. (text) & 33.1 \\
Qwen3-VL-32B-Instruct~\cite{bai2025qwen3} & trajectory (as video) & 21.5 \\ 
Gemini-3-Pro~\cite{comanici2025gemini} & trajectory (as video) & 24.7 \\
SFT Qwen3-VL-4B~\cite{bai2025qwen3} & trajectory (as video) & 23.1 \\
SFT Qwen3-4B-Instruct~\cite{yang2025qwen3} & trajectory (as text) & 25.7 \\
\rowcolor{blue!5}
\ours~embeddings (original) & trajectory & 36.2 \\
\rowcolor{blue!15}
\ours~embeddings (ViPE) & trajectory & \best{46.3} \\
\thickhline
\end{tabular}
\label{tab:dynpose_retrieval}
\end{table}


\begin{figure}[!t]
\centering
\includegraphics[width=0.92\linewidth]{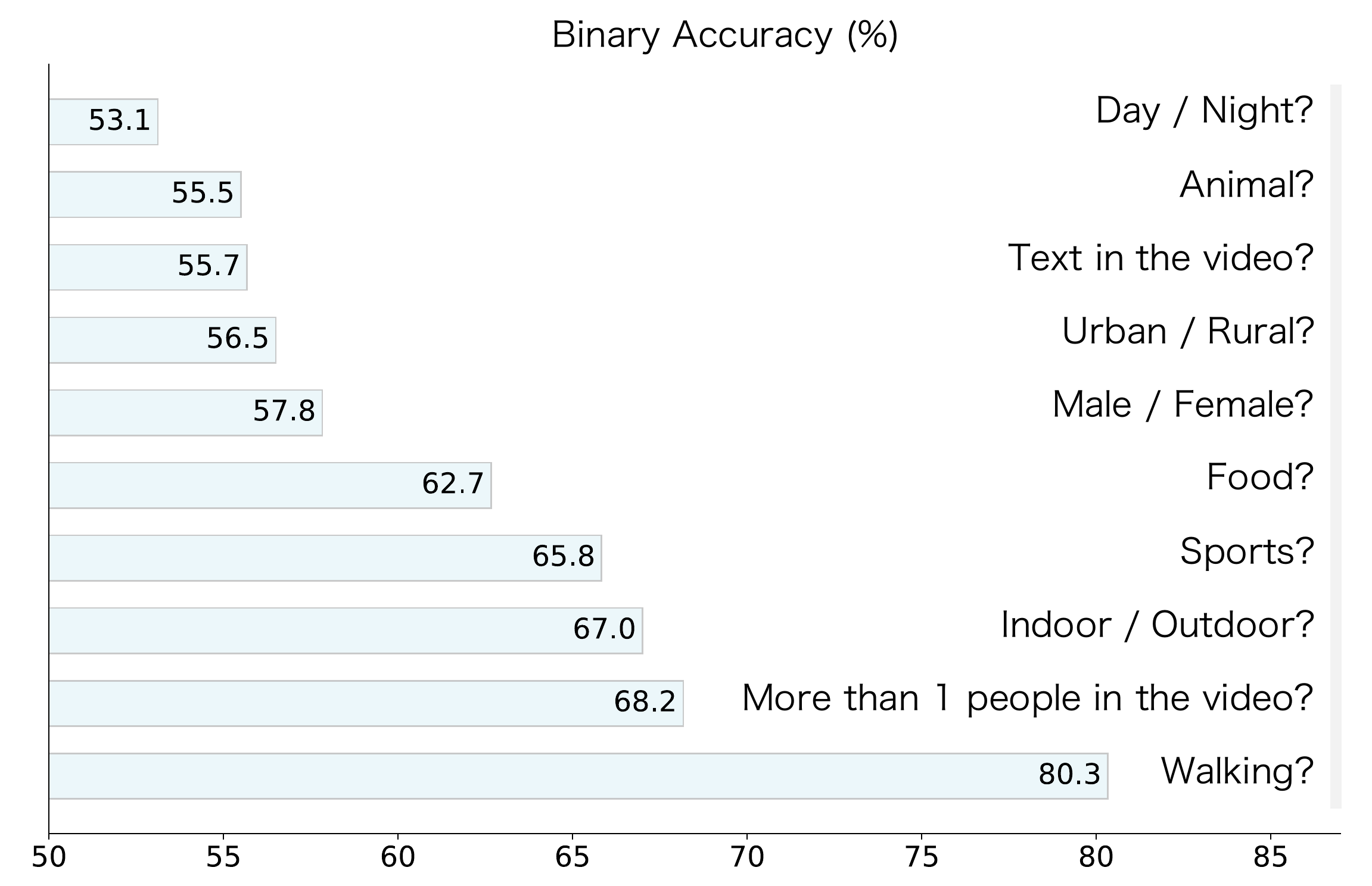}
\vspace*{-3mm}
\caption{\small \textbf{Scene Attribute Classification Results} on DynPose-100K. The results reveal a clear spectrum of what can and cannot be inferred from an observer's camera trajectory.}
\label{fig:dynpose_scene}
\end{figure}

\begin{figure*}[!t]
  \centering
   \includegraphics[width=0.95\linewidth]{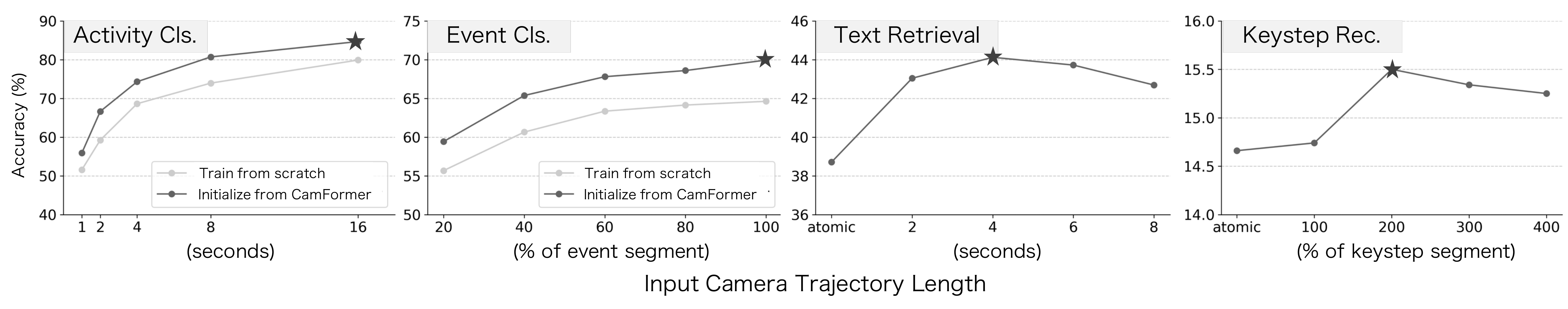}
   \vspace*{-4mm}
   \caption{\small \textbf{Analysis of Contextualized Trajectory Encoding}. For the global label tasks (left two plots), performance monotonically increases with longer input sequences, as more evidence helps disambiguate the activity / event. For the localized label tasks (right two plots), performance peaks at an optimal context length, and then declines as excessive, irrelevant motion acts as noise.
   }
   \label{fig:temporal_context}
\end{figure*}

\begin{figure}[!t]
\centering
\includegraphics[width=0.75\linewidth]{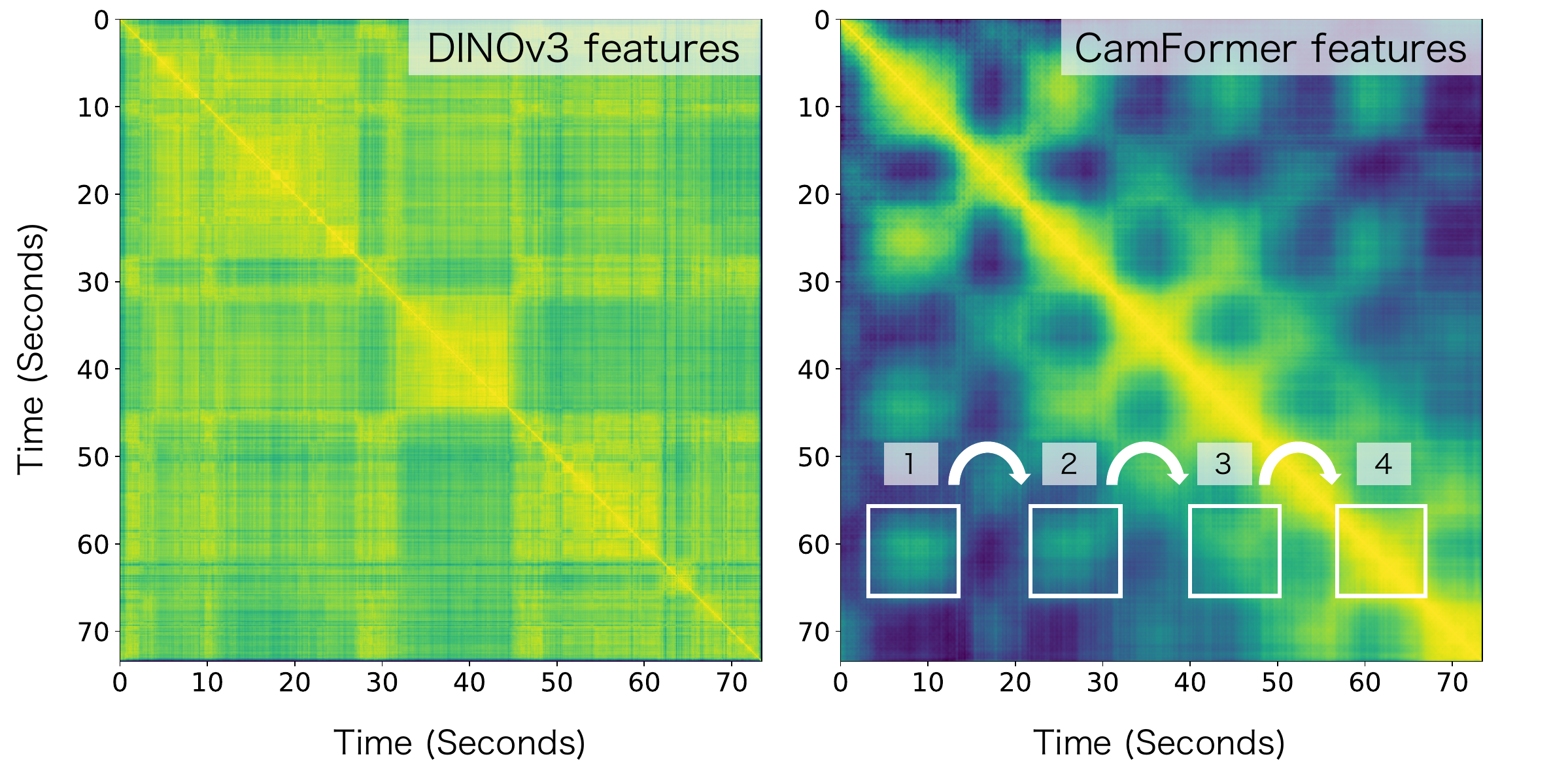}
\vspace*{-3mm}
\caption{\small \textbf{Comparison of Temporal Self-similarity Maps} from {DINOv3~\cite{simeoni2025dinov3} framewise features (left)} and \ours~features (right) on an egocentric cutting sequence from Ego-Exo4D. Our map reveals clear, periodic diagonal streaks that correspond to the number of cuts. The visual feature map produced by DINOv3, in contrast, is unstructured and fails to capture the repeating action. 
}
\label{fig:counting}
\end{figure}

\Q{How robust is \ours\ to estimated poses?} 
\custompar{Evaluation Setup}
Up to now \ours~is pre-trained on the given Aria camera poses~\cite{engel2023project}, which are highly accurate SLAM-based multi-sensor estimates. 
We further investigate the robustness of \ours~on more accessible, video-based camera pose estimates.
We select two representative tasks on Ego-Exo4D for this analysis: coarse-grained activity classification, which we evaluate with end-to-end fine-tuning, and fine-grained keystep recognition, which we evaluate using a linear SVM on frozen features.

\custompar{Results}
Table~\ref{tab:egoexo4d_camest_combined} shows that our \ours, pre-trained only on Aria trajectories, is robust and generalizes well across three different estimated poses. First, for activity classification, the benefit of our pre-training holds: a model initialized with our Aria-pretrained \ours~remains superior to one trained from scratch, even when evaluated on these video-based camera estimates. Second, for keystep recognition, the trajectory's complementary power persists, as the video-trajectory fusion approach (using estimated poses) continues to outperform the video-only baseline. 

Among the estimators, $\pi^3$~\cite{wang2025pi} emerges as the top performer. Yet, there are still gaps of these video-only pose estimation methods to Aria trajectories, which highlight the need for further improvements in estimation accuracy. 
Our analysis also introduces a valuable new direction: a novel semantic benchmark that evaluates estimators on their downstream semantic utility, offering a more holistic assessment than geometric error alone.

\subsection{\ours\ for Exocentric Videos}\label{sec:exp_results_exo}
\Q{Does the trajectory-semantic link exist in exocentric videos (third-person view)?}

\noindent {Building on our egocentric findings, we now analyze the exocentric domain, where we posit the attentional imprint of the observer is a rich signal for the perceived event. For instance, the quick, reactive pans of an observer filming a soccer match are likely to differ from the slow, steady hold they would use to film a museum painting. }
\custompar{Evaluation Setup} We evaluate the exocentric \ours~(trained on DynPose-100K) on four tasks: (1) text retrieval, where we use the same 5-way MCQ setup as in the egocentric domain, on a 1000-query DynPose-100K test set. \CR{We evaluate against three classes of strong LMM baselines. See Supp. for a detailed description.}
(2) scene attribute classification, where we design a list of 10 attribute questions and automatically label a held-out set on DynPose-100K via LMM prompting; (3) event classification on FineGym using its 4 event labels; and (4) action recognition on UCF-Dynamic, our custom 8-class subset of UCF101. 

\custompar{Results}
\begin{itemize}
    \item \textit{Text Retrieval}. Our exocentric results (Table~\ref{tab:dynpose_retrieval}) confirm that a meaningful link between observer trajectories and video content exists. Our \ours~performs well above random chance (20\%) across both pose sources (ViPE and the original). 
    Importantly, it outperforms 
    \CR{all two-stage, zero-shot and finetuned LMM} baselines. This demonstrates the effectiveness of interpreting camera trajectories directly and suggests that current LMMs have yet to perfect the generation of accurate camera motion descriptions. 
    Fig.~\ref{fig:qualitative_retrieval} (bottom) shows one qualitative example, where our \ours~correctly links a circling overview motion to its corresponding scene description.
    \item \textit{Scene Attribute Classification}. Echoing the activity-level analysis in the egocentric domain, we test our exocentric \ours's ability to classify a range of binary scene attributes on DynPose-100K (Fig.~\ref{fig:dynpose_scene}), inspired by~\cite{wang2025seeing}. The analysis reveals what can be inferred from an observer's camera motion. At one end, it provides no discernible signal for static attributes like day vs. night, where performance is near random. At the other end, it is a strong predictor for physical attributes, \eg, achieving over 80\% accuracy for classifying walking behavior.
    \item \textit{Event \& Action Classification}. As summarized in Fig.~\ref{fig:results_overview} (a), our \ours~learned on DynPose-100K, effectively delivers benefits to semantic classification tasks. Compared to a base model trained from scratch, initializing from \ours~achieves a +5.3\% gain on FineGym event classification and a +4.0\% gain on UCF-Dynamic action recognition. See Supp. for full results. 
\end{itemize}



\subsection{Further Analysis}\label{sec:exp_results_analysis}

\paragraph{Analysis of Temporal Context Length.} We validate our contextualized trajectory encoding strategy (Fig.~\ref{fig:temporal_context}) by varying the input camera trajectory length across two settings:
\begin{itemize}
    \item For \textit{tasks with one global label} (Ego-Exo4D activity and FineGym event classification), the left two plots show that performance steadily improves with longer sequences, as more evidence helps disambiguate the overall activity / event. Furthermore, our model initialized from the pre-trained \ours~consistently outperforms the train-from-scratch baseline across all sequence lengths, showcasing the benefits of our proposed pre-training. 
    \item For \textit{tasks with temporally localized labels} (Ego-Exo4D text retrieval and keystep recognition), we investigate the impact of including temporal context from outside the labeled time window. The right two plots show that performance initially improves as more context is added, demonstrating that surrounding motion is crucial for understanding an action. This trend reverses when the context window becomes excessively large, as irrelevant motion begins to act as noise. This reveals an optimal context ``sweet spot" for these localized tasks.
\end{itemize}


\para{Repetitive Action Counting} is an interesting emergent capability of \ours. To do this, we compute the self-similarity map of \ours's temporal features (\ie, the output token sequence before the final average pooling). As shown in Fig.~\ref{fig:counting}, our map (right) reveals clear, periodic diagonal streaks for an egocentric cutting sequence, in contrast to the unstructured map from DINOv3~\cite{simeoni2025dinov3} visual features (left). These periodic patterns are precisely the type of fine-grained temporal signal required for counting~\cite{dwibedi2020counting}. 
See Supp. for more qualitatives and an animation.


\section{Conclusion}


This work challenges the traditional, geometric view of camera trajectory, showcasing its potential as a semantic signal. 
Our results with \ours~reveal the trajectory-semantic link exists in both egocentric and exocentric domains. 
Crucially, \ours~is robust across diverse pose {estimation methods}, demonstrating its practical utility for real-world videos. We hope these findings inspire the community to take a new look at the camera trajectory and further explore this promising modality.

\section*{Acknowledgment}
UT Austin is supported in part by the IFML NSF AI Institute.  We thank Junyu Xie for assistance and valuable suggestions on the camera pose estimation pipelines; Jean-Baptiste Alayrac, Carl Doersch and Ignacio Rocco for constructive feedback; and Ang Cao, Yue Zhao and Lorenzo Torresani for helpful discussions.

{
    \small
    \bibliographystyle{ieeenat_fullname}
    \bibliography{main}
}

\clearpage
\setcounter{page}{1}
\setcounter{section}{0}
\maketitlesupplementary

\section{Supplementary Video}
We invite readers to view the supplementary video 
available at \url{https://sites.google.com/view/seeing-without-pixels}
for a visual demonstration of our work's overview and additional qualitative examples. The video animates the static trajectories presented in the paper (\eg, Fig.~\ref{fig:teaser}), offering a clearer view of the motion signatures. It also visualizes the learned embedding space, demonstrating how \ours~clusters semantically similar neighbors. Furthermore, we provide qualitative examples of successful text retrieval (in both egocentric and exocentric domains) and demonstrate \ours's emergent capabilities, such as repetitive action counting and text-based trajectory retrieval, alongside a visual analysis of failure modes.

\section{Experimental Setup}
\subsection{Task Setup}
\subsubsection{Text Retrieval (5-way MCQ)} To evaluate cross-modal alignment between the camera trajectory and text modality, we formulate text retrieval as a 5-way MCQ task, mitigating the inherent ambiguity of open-ended retrieval. Given a trajectory query, the model must select the correct text description from five options based on feature similarity, using the pre-trained \ours~as a frozen feature extractor.

\paragraph{Negative Sampling Strategy.} To ensure a rigorous benchmark, we curate distractor options to have primary action verbs that do not overlap with the ground truth. Furthermore, we adopt a hard negative sampling strategy: distractor options are sourced from narrations within the same continuous video take (for egocentric datasets) or from captions with the same YouTube ID (for exocentric datasets), forcing the model to distinguish between temporally or thematically adjacent actions.

\paragraph{Evaluation Splits.} This task serves as our primary testbed across Ego-Exo4D, Nymeria, and DynPose-100K. We further exploit dataset-specific annotations to dissect our model's strengths versus the visual modality. On Ego-Exo4D, we analyze performance across ``in-view'' and ``out-of-view'' splits. On Nymeria, we break down results by narration type (legs, focus, body, hands).Qualitative MCQ examples are provided in Fig.~\ref{fig:qualitative_retrieval} (main paper) and Fig.~\ref{fig:more_qual}. 

\subsubsection{Other Evaluation Tasks}
\paragraph{Proficiency Estimation on Ego-Exo4D.} We utilize the dataset's skill-level annotations for the rock climbing and music activities. We focus exclusively on these two scenarios as they are the only ones that retain sufficient samples and a balanced class distribution after filtering for camera pose availability. To further mitigate imbalance, we formulate the task as a binary classification problem (expert vs. non-expert). We evaluate using an end-to-end fine-tuning protocol, comparing our pre-trained \ours~initialization against a model trained from scratch.

\paragraph{Keystep Recognition \& Localization on Ego-Exo4D.} We utilize the dataset's established benchmark, which defines 278 fine-grained keystep labels. We adopt a frozen feature evaluation protocol for both tasks: we train a linear SVM for recognition and a dedicated localization network~\cite{mu2024snag} for localization, both on top of our fixed trajectory embeddings.

\paragraph{Activity Classification on Ego-Exo4D.} We evaluate coarse-grained understanding using the dataset's 8 high-level activity labels. For this task, we adopt an end-to-end fine-tuning protocol, training the trajectory encoder jointly with a linear classification head. We compare initializing from our pre-trained \ours~against a model trained from scratch to measure the benefit of pre-training.

\paragraph{Scene Attribute Classification on DynPose-100K.}
We enrich DynPose-100K with semantic labels to enable a new binary scene attribute classification task. To do this, we designed 10 questions covering a diverse range of attributes (e.g., temporal, environmental, and social context) and utilized Gemini-2.5-Pro~\cite{comanici2025gemini} to automatically label the videos. The prompts used for annotation are as follows:
\vspace*{2mm}
\begin{enumerate}
    \item \textbf{Day / Night:} Is the video filmed during the day or at night?
    \item \textbf{Animal:} Does the video contain any animals?
    \item \textbf{Text:} Does the video contain any visible written text?
    \item \textbf{Urban / Rural:} Is the scene in the video urban or rural?
    \item \textbf{Male / Female:} What is the gender of the people in the video?
    \item \textbf{Food:} Does the video feature any food?
    \item \textbf{Sports:} Is the video related to sports activities?
    \item \textbf{Indoor / Outdoor:} Decide if the video is filmed indoors, outdoors, or if it is unclear.
    \item \textbf{More than 1 people:} How many people are visible in the video?
    \item \textbf{Walking:} Does the video show people walking?
\end{enumerate}
\vspace*{2mm}
From these new labels, we create a balanced 3,000-sample dataset for each attribute (with equal positive and negative examples) and train a linear SVM on our frozen camera trajectory features using an 80:20 train/test split.

\paragraph{Event Classification on FineGym.} We evaluate on the 4 gymnasium event labels: floor exercise, balance beam, uneven bars, and vault. We adopt an end-to-end fine-tuning protocol, training the encoder jointly with a linear classification head.

\paragraph{Action Recognition on UCF101-Dynamic.} To ensure a meaningful, motion-based evaluation, we curated this custom benchmark by quantitatively analyzing camera dynamics in UCF101. We selected the two most dynamic classes from each of four major action types, resulting in 8 classes: SkiJewandb sync ./anonymized/runs --project test-privacyt, SkateBoarding, Knitting, MoppingFloor, WalkingWithDog, Lunges, MilitaryParade, and SoccerPenalty. We evaluate using the same end-to-end fine-tuning protocol. For completeness, we also report results on the full 101-class UCF101 dataset (cf. Table~\ref{tab:ucf101}).

\subsection{Datasets}
\paragraph{Pretraining Data.} For the egocentric domain, we use Ego-Exo4D~\cite{grauman2024ego}. Adhering to the official splits, we obtain 159,186 training and 69,073 validation (trajectory, text) pairs, where text is human-annotated narrations provided by the dataset. Trajectory boundaries for these long, untrimmed videos are defined following~\cite{lin2022egocentric,pramanick2023egovlpv2}. There are two camera trajectory sources: the original Aria glasses data (which we downsample to 20 FPS) and video-only estimations we obtain by running $\pi^3$ (5 FPS); a comparison is provided in Table~\ref{tab:pretraining_ablation}. For the exocentric domain, we use DynPose-100K~\cite{rockwell2025dynamic}. As no official split is available, we randomly split the dataset into 88,151 training and 10,452 validation (trajectory, text) pairs, where the text is video captions from Panda70M~\cite{chen2024panda}. Unlike the egocentric data, these are short clips with fixed boundaries. There are two trajectory sources, both estimated from videos: the original dataset's provided ones (12 FPS) and the ViPE-provided~\cite{huang2025vipe} ones (30 FPS).

\paragraph{Downstream Data.} For the egocentric domain, we evaluate on Ego-Exo4D and Nymeria. On Ego-Exo4D, for our designed text retrieval task, we draw samples from the official validation split; for all other tasks, we follow the dataset's official benchmark splits. For Nymeria, which shares the same Aria hardware as Ego-Exo4D, we use the entire set of data with available motion narrations as a zero-shot test set for text retrieval; the camera trajectory is also downsampled to 20FPS. For the exocentric domain, we use the original action labels from FineGym and UCF101. Since these datasets lack trajectories, we generate them using our pose estimation pipeline at 5 FPS. We generate three versions for UCF101 (using MegaSaM~\cite{li2025megasam}, ViPE~\cite{huang2025vipe}, and $\pi^3$~\cite{wang2025pi}) and one version for FineGym (using $\pi^3$).

\subsection{Implementation}
When using hardware-estimated camera poses (\ie, from Aria glasses~\cite{engel2023project}), we utilize the gravity direction information. To be specific, we compute the 3D gravity vector in our chosen relative reference frame and project it to $d_{in}$ via a learned linear projection layer. This additional token is subsequently prepended to the input sequence before processing by the Transformer. This step is omitted when processing poses estimated from monocular videos.
We train \ours~using an AdamW optimizer with a learning rate of $1 \times 10^{-4}$, weight decay of $1 \times 10^{-3}$, and a batch size of 1024. CLIP loss temperature $\tau$ is 0.07. Training is conducted on 8 NVIDIA A100 GPUs. 


\paragraph{Camera Pose Extraction.} 
Given the computational expense of running multiple pose estimators, all pose estimations are performed at 5 FPS. For the particularly long, untrimmed videos in the Ego-Exo4D activity classification task (which can be several minutes), we further limit pose extraction to the center 4-second clip. For the egocentric setting, an alignment step is required. Our model is pre-trained on the Aria pose coordinate frame (x left, y up, z forward), so we apply a rigid transformation to convert all estimated poses from the standard OpenCV frame (x right, y down, z forward) before they are fed into our model.

\begin{figure*}[!t]
  \centering
  \includegraphics[width=1.0\linewidth]{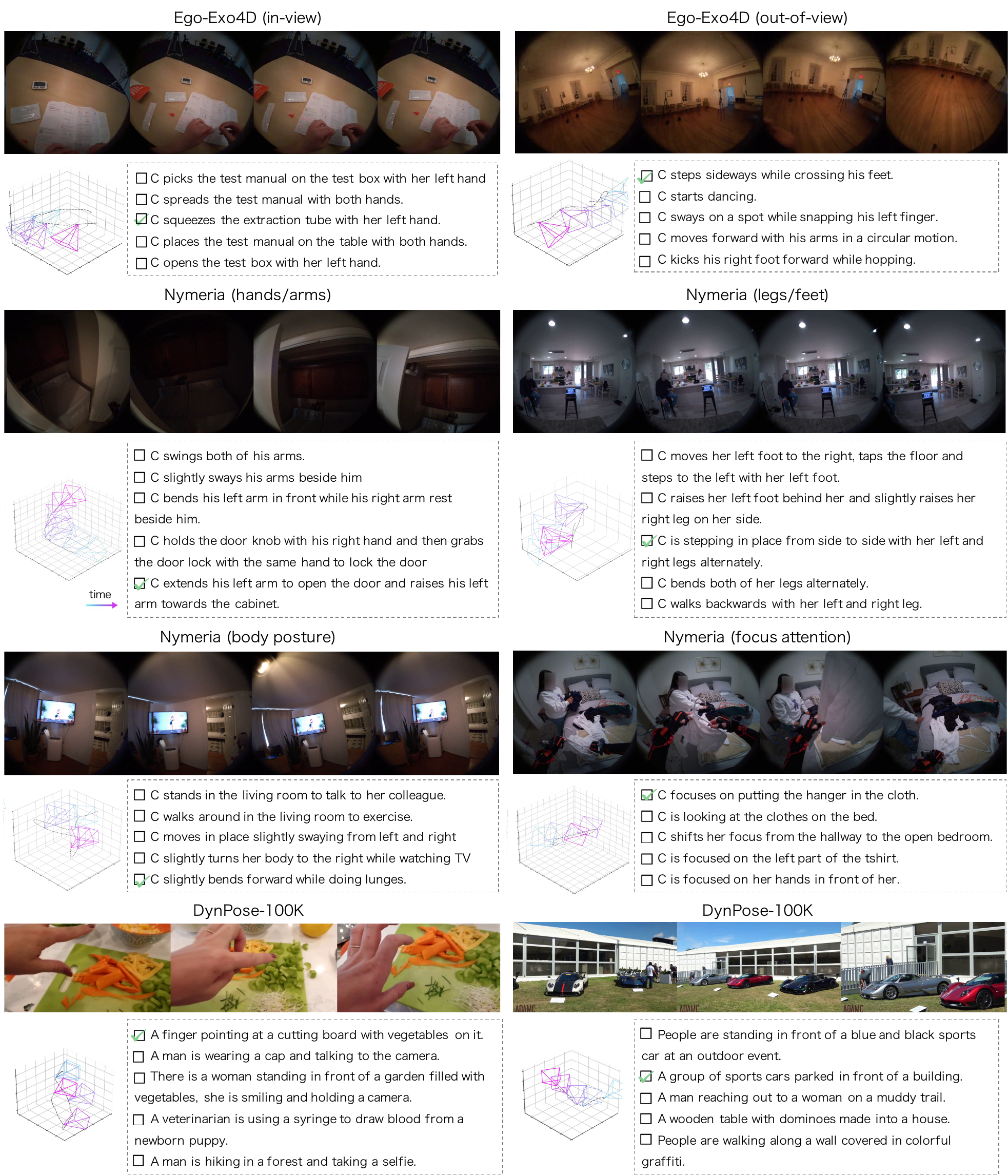}
    \caption{Qualitative Text Retrieval Results on Ego-Exo4D (top row), Nymeria (middle two rows) and DynPose-100K (bottom row). Note that \ours~takes only the camera trajectory as input; corresponding video frames are shown solely for illustration. These examples further demonstrate that our model effectively captures the trajectory-semantic link across both egocentric and exocentric domains.}
    \label{fig:more_qual}
\end{figure*}

\begin{figure}[!t]
  \centering
  \includegraphics[width=0.9\linewidth]{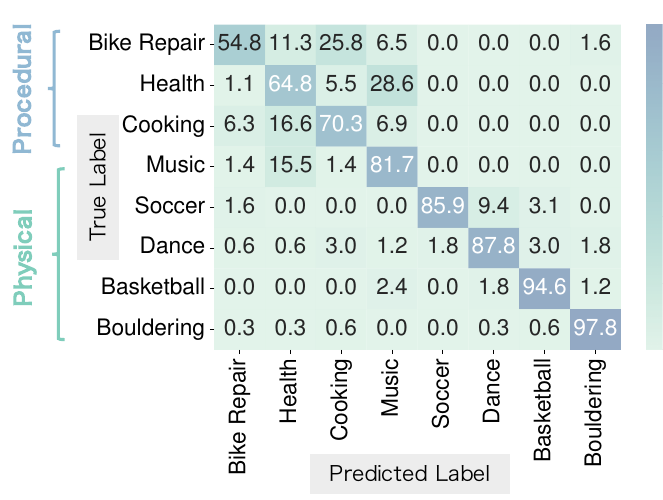}
    \caption{Confusion Matrix for \ours~on Ego-Exo4D Activity Classification. \ours~performs strongly on dynamic physical activities (\eg, near-perfect for ``bouldering''), while the main confusion occurs between the three procedural activities, which involve more subtle motion cues.}
    \label{fig:scenario_cls_confusion}
\end{figure}

\begin{figure}[!t]
  \centering
  \includegraphics[width=0.9\linewidth]{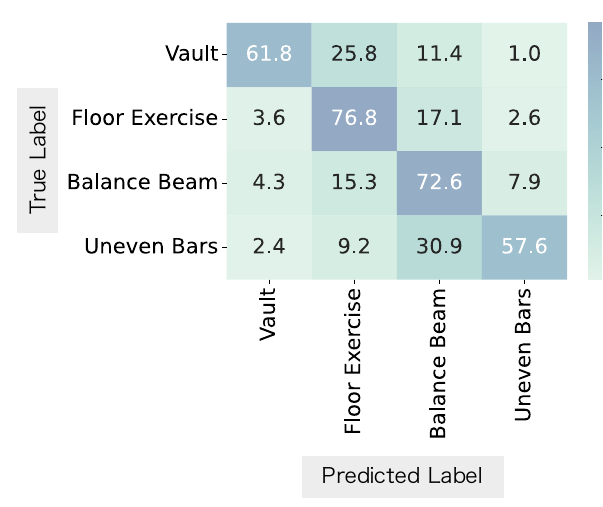}
    \caption{Confusion Matrix for \ours~on FineGym Event Classification. The matrix details our model's performance on the 4 gymnasium activities.}
    \label{fig:finegym_confusion}
\end{figure}

\begin{figure*}[!t]
  \centering
  \includegraphics[width=1.0\linewidth]{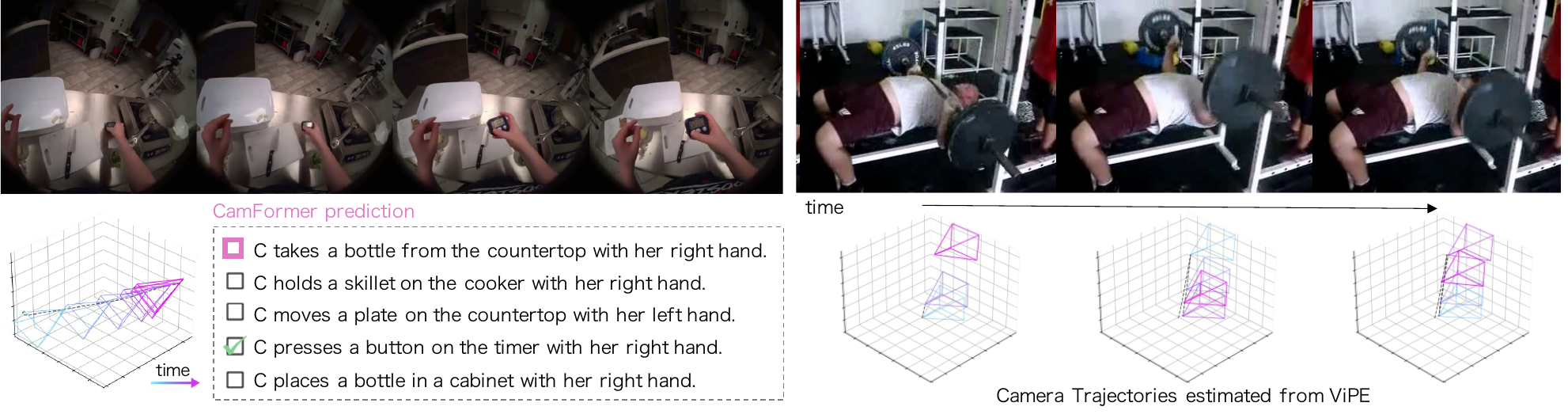}
    \caption{Failure Cases. Left: \ours~struggles to distinguish actions with subtle motion patterns and often cannot capture specific noun semantics. Right: The pose source fails when the ViPE estimator mistakenly tracks object motion as camera motion (best viewed in Supp. video), highlighting a necessary area for development in pose estimation algorithms.}
    \label{fig:failure_case}
\end{figure*}

\paragraph{Analysis of Contextualized Trajectory Encoding.} We detail the experimental setup for the four settings in Fig.~\ref{fig:temporal_context} of the main paper.
\begin{itemize}
    \item \emph{Global Label Tasks}. We fine-tune \ours~end-to-end for these tasks. Our model accepts flexible input trajectory lengths (both in pre-training and finetuning), which allows us to systematically vary the input length during inference. (1) For Ego-Exo4D activity classification, we vary the input trajectory length from 1 to 16 seconds. (2) For FineGym event classification, since events have fixed segments, we vary the input ratio from 20\% to 100\% of the full event duration.
    \item \emph{Localized Label Tasks}. For these tasks, we apply the pre-trained \ours~as a frozen feature extractor and investigate extending the context outside the given segment window $[t_1, t_2]$. (3) For Ego-Exo4D text retrieval, we extend the atomic window $[t_1, t_2]$ by a total duration of 0, 2, 4, 6, or 8 seconds. We apply this symmetrically; for instance, a 2-second extension results in the input window $[t_1-1, t_2+1]$. (4) For Ego-Exo4D keystep recognition, we expand the input trajectory window proportionally (100\%-400\% of the original duration).
\end{itemize}

\subsection{Baselines}
For the Gemini row in Table~\ref{tab:main_retrieval_results} of the main paper, we input 8 uniformly sampled video frames directly. The five candidate texts are randomly shuffled and assigned labels (A-E). Gemini-2.5-Pro~\cite{comanici2025gemini} is queried with the prompt: \textit{Which of the following descriptions best matches the video?}

\noindent
For exocentric text retrieval (Table~\ref{tab:dynpose_retrieval} of the main paper), we consider the following baselines:
\begin{itemize}
    \item \textbf{Two-stage camera description baselines.} First, we prompt specialized LMMs (Qwen-VL-7B~\cite{fang2025camerabench} or ShotVL-7B~\cite{liu2025shotbench}) to \textit{Describe the camera motion in this video.} Note that these models are trained to generate textual description of the camera motion in the associated video (\eg, ``zoom'', ``pan'') and do not aim to describe the content of the video. Second, we feed this generated description into another LLM (we use Gemini-2.5-Flash~\cite{comanici2025gemini}) and prompt it to answer the MCQ given the camera motion described in text form. The prompt is: \textit{The following describes the motion and focus of a camera while filming a scene:[Generated Description]. Which of the following events or scene descriptions is most likely being filmed with this camera movement? [Option A-E].}
    \CR{\item \textbf{Zero-shot LMM baselines.} We evaluate the zero-shot capabilities of two strong LMMs for this task: Qwen3-VL-32B-Instruct~\cite{bai2025qwen3} (open-source) and Gemini-3-Pro~\cite{comanici2025gemini} (proprietary). To make the trajectory data compatible with these models, we render the 3D camera trajectories into video clips, consistent with our human-readable visualizations. We provide these trajectory videos as input alongside the following prompt: \textit{The video features the trajectory of a camera. Answer the question about the video content based on the camera trajectory. Which of the following events or scene descriptions is most likely being filmed with this camera movement? [Option A-E].}
    \item \textbf{SFT baselines.} To assess the impact of domain-specific training, we implement two supervised fine-tuning (SFT) baselines using Qwen3-4B-Instruct~\cite{yang2025qwen3} (an LLM) and Qwen3-VL-4B-Instruct~\cite{bai2025qwen3} (an LMM). We perform SFT on the DynPose-100K training set to predict video descriptions. For the LMM, we input the rendered trajectory videos and the zero-shot prompt above. For the text-only LLM, we format the trajectory as a text list of 9D pose sequences with the following prompt: \textit{You are given the relative camera trajectory of a video as a sequence of pose vectors. Each pose vector is 9D: 3D translation + 6D rotation. Camera trajectory: [value]. Task: Which of the following events or scene descriptions is most likely being filmed with this camera movement? [Option A-E].}}
\end{itemize}

\section{Results}
\subsection{Additional Results}
\begin{table*}[!t]
\tablestyle{4pt}{1.2}
\caption{Per-Activity Breakdown of Ego-Exo4D Text Retrieval Results (MCQ Accuracy). We compare the performance of \ours~(trajectory features) and the best video baseline, EgoVLPv2~\cite{pramanick2023egovlpv2} (Ego-Exo4D), across all 8 activity scenarios. This detailed breakdown allows us to identify the relative strengths and weaknesses of the camera trajectory versus the video modality.}
\begin{tabular}{lccccccccccccccccccccccc}
\thickhline
 & \multicolumn{2}{c}{{Bike}} & \multicolumn{2}{c}{{Health}} & \multicolumn{2}{c}{{Cooking}} & \multicolumn{2}{c}{{Music}} & \multicolumn{2}{c}{{Soccer}} & \multicolumn{2}{c}{{Dance}} & \multicolumn{2}{c}{{Basketball}} & \multicolumn{2}{c}{{Bouldering}} & \multirow{2}{*}{All}\\
& iv & oov & iv & oov & iv & oov & iv & oov & iv & oov & iv & oov & iv & oov & iv & oov & & \\
\hline
\# Queries & 500 & 205 & 500 & 416 & 500 & 500 & 500 & 176 & 500 & 500 & 282 & 500 & 500 & 500 & 500 & 500 & 7079 \\
Video & \best{37.20} & \best{40.98} & \best{52.60} & \best{47.12} & \best{61.60} & \best{45.80} & 29.40 & 21.02 & 24.40 & 21.20 & 25.89 & 15.40 & 50.80 & 38.00 & 59.40 & 29.60 & 38.40 \\
Ours & 28.80 & 28.78 & 24.40 & 26.44 & 49.80 & 39.40 & \best{32.00} & \best{35.23} & \best{52.00} & \best{26.60} & \best{39.36} & \best{50.60} & \best{67.40} & \best{57.00} & \best{62.60} & \best{55.40} & \best{44.80} \\
\thickhline
\end{tabular}
\label{tab:egoexo4d_retrieval_peractivity}
\end{table*}

\paragraph{Text Retrieval.} Supplementing Table~\ref{tab:main_retrieval_results} of the main paper, Table~\ref{tab:egoexo4d_retrieval_peractivity} provides a detailed activity-level breakdown on Ego-Exo4D text retrieval, comparing our \ours~embeddings with leading video encoder features (EgoVLPv2~\cite{pramanick2023egovlpv2}). The results allow us to clearly delineate the strengths of the two modalities. For the procedural activities (where visual cues are more critical), the video baseline maintains its lead. For the five physical activities, the camera trajectory modality is demonstrably stronger on its own. This performance is particularly effective in out-of-view (oov) settings, highlighting trajectory's unique value in scenarios where the visual signal is occluded or ambiguous. Finally, our \ours~achieves the best overall result, demonstrating the promising value of the camera trajectory in action understanding.

\paragraph{Qualitative Results.} Supplementing Fig.~\ref{fig:qualitative_retrieval} of the main paper, Fig.~\ref{fig:more_qual} presents additional qualitative results for the text retrieval task. These examples demonstrate that across diverse datasets and varying text descriptions, \ours~ successfully decodes the semantic information embedded in camera trajectories and accurately matches it with the corresponding text.

\paragraph{Proficiency Estimation.} For physical activities, we find that camera trajectory is a powerful standalone signal for assessing skill levels (beginner/expert). Table~\ref{tab:egoexo4d_proficiency} presents proficiency estimation results for the two physical activities. Our lightweight model successfully captures the motion signatures of expertise and outperforms the video baseline. This performance is further boosted by our pre-training strategy, which surpasses the train-from-scratch counterpart.
\begin{table}[!t]
\tablestyle{6pt}{1.2}
\caption{Proficiency Estimation Accuracy (\%) on Ego-Exo4D. The camera trajectory is a particularly strong modality for this physical task, outperforming the video baseline. Moreover, our pretraining is crucial, as initializing from \ours~provides a boost over training from scratch. }
\begin{tabular}{lcccc}
\thickhline
Method & Modality & Pretrain? & Bouldering & Dancing\\
\hline
Majority & - & - & 55.97 & 59.30 \\
TimeSformer~\cite{bertasius2021space} & video & - & 55.35 & 69.92 \\
\rowcolor{blue!5}
\ours & trajectory & \ding{55} & 63.52 & 66.67 \\
\rowcolor{blue!15}
\ours & trajectory & \ding{51} & \best{65.41} & \best{70.73} \\
\thickhline
\end{tabular}
\label{tab:egoexo4d_proficiency}
\end{table}


\paragraph{Keystep Recognition \& Localization.} 
\begin{table}[!t]
\tablestyle{3pt}{1.2}
\caption{Keystep Recognition and Localization Results on Ego-Exo4D. For these procedural tasks, where vision is a strong baseline, fusing trajectory and video features (denoted by $^{\star}$) consistently outperforms the video-only model, proving that camera trajectory provides an essential, non-redundant signal.}
\begin{tabular}{lcccccc}
\thickhline
\multirow{3}{*}{Method} & \multirow{3}{*}{Modality} & Rec. & \multicolumn{2}{c}{Loc. Rank@1}  & \multicolumn{2}{c}{Loc. Rank@5} \\
 & & Acc. & IoU & IoU & IoU & IoU \\
 & & (\%) & @0.3& @0.5 & @0.3 & @0.5 \\
\hline
Majority & - & 3.52 & - & - & - & -\\
EgoVLPv2~\cite{pramanick2023egovlpv2} & video & 29.17 & 31.81 & 26.28 & 62.90 & 52.69\\
\rowcolor{blue!5}
\ours & trajectory & 14.07 & 20.29 & 15.67 & 47.23 & 38.09\\
\rowcolor{blue!15}
\ours$^{\star}$ & video+trajectory & \best{32.37} & \best{34.68} & \best{29.06} & \best{66.65} & \best{57.29}\\
\thickhline
\end{tabular}
\label{tab:egoexo4d_keystep}
\end{table}

 For procedural activities, where motion patterns can be ambiguous, camera trajectory provides valuable complementary information. For the keystep recognition and localization tasks on these scenarios (Table~\ref{tab:egoexo4d_keystep}), fusing trajectory with video features provides a consistent performance boost over the video-only baseline.

\paragraph{Activity \& Event Classification.} The confusion matrix (Fig.~\ref{fig:scenario_cls_confusion}) on Ego-Exo4D activity classification provides a detailed breakdown of the per-class accuracy results in Fig.~\ref{fig:results_overview} (c). \ours~excels at recognizing dynamic physical activities, while confusion is heavily concentrated among the three procedural activities (where camera motion is subtle). For the exocentric domain, on FineGym event classification (Fig.~\ref{fig:finegym_confusion}), the model performs strongly on recognizing `floor exercise', and the main confusion occurs between `uneven bars' and `balance beam'. 


\begin{table}[!t]
\tablestyle{2pt}{1.2}
\caption{Comparing action recognition results on UCF101-Dynamic (left) and UCF101 (right) with various estimated camera poses. Echoing our egocentric analysis, the results confirm the benefits of our pre-training strategy. Across all pose estimators, the model initialized with our checkpoint pre-trained on DynPose-100K (\ding{51}) consistently outperforms its counterpart trained from scratch (\ding{55}).}
\begin{tabular}{lcccccc}
\thickhline
\multirow{2}{*}{Pose Source} & \multicolumn{3}{c}{UCF101-Dynamic} & \multicolumn{3}{c}{UCF101} \\
 & Pretrain \ding{55} & Pretrain \ding{51} & $\Delta$ & Pretrain \ding{55} & Pretrain \ding{51} & $\Delta$ \\
\hline
MegaSaM~\cite{li2025megasam} & 66.67 & 69.15 & \gain{2.48} & 16.02 & 17.91 & \gain{1.89} \\
ViPE~\cite{huang2025vipe}    & 64.18 & 68.16 & \gain{3.98} & 16.54 & 19.62 & \gain{3.08} \\
$\pi^3$~\cite{wang2025pi}     & 61.69 & 64.18 & \gain{2.49} & 17.53 & 19.25 & \gain{1.72} \\
\thickhline
\end{tabular}
\label{tab:ucf101}
\end{table}
\paragraph{Action Recognition.} Table~\ref{tab:ucf101} compares action recognition performance using our pre-trained \ours~against the train-from-scratch baseline on UCF-Dynamic (our curated 8-class subset) and the full UCF101 dataset. The results show that initializing from \ours~yields better performance than the train-from-scratch baseline across all three pose sources and on both settings. The largest performance gain is observed when using ViPE poses, as \ours~was pre-trained with ViPE camera trajectories on DynPose-100K. Even with the other pose sources, the consistent gains observed across datasets demonstrate the robust generalization capability of our pre-training strategy.

\begin{table}[!t]
\tablestyle{6pt}{1.2}
\caption{
Ablation Study of Input Camera Trajectory Representation on Ego-Exo4D Text Retrieval. We compare various formulations, including the use of absolute vs.\ relative poses, the rotation format (no vs. 4D quaternion vs. 6D continuous), the specific reference frame used for calculating relative poses, and whether to include gravity direction information.} 
\begin{tabular}{lcc}
\thickhline
 & Dim. (Tsl. + Rot.) & Acc. (\%) \\
\hline
Absolute & 3D & 32.84 \\
Absolute & 3D + 4D & 34.74 \\
Absolute & 3D + 6D & 37.82 \\
\hdashline
Relative (prev.) & 3D + 4D & 43.66 \\
Relative (mid.) & 3D + 4D & 44.02 \\
Relative (any) & 3D + 4D & 44.00 \\
\hdashline
Relative (mid.) & 3D + 6D & 44.12 \\
+ Gravity direction & 3D + 6D & \best{44.81} \\
\thickhline
\end{tabular}
\label{tab:campose_ablation}
\end{table}
\begin{table}[!t]
\tablestyle{6pt}{1.2}
\caption{Ablation Study of Pretraining Choices on Ego-Exo4D text Retrieval. We compare \ours~performance using two different camera pose sources (high-fidelity Aria~\cite{engel2023project} vs. video-estimated $\pi^3$~\cite{wang2025pi}) and text encoder training modes (frozen vs. finetuned).} 
\begin{tabular}{lccc}
\thickhline
& Pose Source & Text Encoder & Acc. (\%) \\
\hline
(a) & $\pi^3$~\cite{wang2025pi} & frozen & 43.86 \\
(b) & Aria~\cite{engel2023project} & finetune & 45.42 \\
(c) & Aria~\cite{engel2023project} & frozen & 44.81 \\
\thickhline
\end{tabular}
\label{tab:pretraining_ablation}
\end{table}

\CR{\paragraph{Semantic Disentanglement in Embedding Space.} To confirm that \ours~learns generalized semantic representations rather than relying on geometric similarity, we analyze 5000 validation samples from Ego-Exo4D. We identify the top 1\% most similar trajectory pairs in the raw input space, as measured by Euclidean distance after trajectory alignment. When mapped to the \ours~embedding space, only 1.61\% of these pairs remain in the top 1\% of similarity, demonstrating that the model actively separates geometrically similar inputs. Crucially, this disentanglement is semantically driven: the pairs that remained close in our embedding space exhibited a 65.8\% agreement in their activity labels, compared to only a 7.6\% agreement for pairs that the model separated.}

\subsection{Ablation Study} Table~\ref{tab:campose_ablation} presents the ablation study on Ego-Exo4D text retrieval, where we compare various ways to represent the input camera trajectory. The results demonstrate that relative pose sequences are critical and greatly outperform absolute pose sequences, with the sequence midpoint being the optimal reference frame. Furthermore, the 6D continuous rotation representation~\cite{zhou2019continuity} is preferred over the 4D quaternion, and encoding gravity direction provides a further performance boost. 

Table~\ref{tab:pretraining_ablation} investigates our pretraining choices. First, regarding pose source: replacing high-fidelity Aria poses with video-estimated $\pi^3$ ones still yields comparable performance (43.86\% vs. 44.81\%). This is a promising result, indicating significant potential to scale up pre-training data using poses estimated from large collections of in-the-wild videos. Second, regarding the text encoder: while fine-tuning the CLIP encoder yields a marginal performance gain, it comes with a substantial computational cost. We therefore adopt the frozen text encoder for our final model to prioritize efficiency, though we posit that end-to-end fine-tuning may become more beneficial as data scale increases in the future.

\subsection{Limitations} \label{supp_sec:limitations}

We acknowledge that obtaining high-quality camera poses initially incurs a computational cost, whether through multi-sensor hardware or video estimation algorithms. We view this, however, as a one-time, amortized process. Concurrent advances in hardware and the development of efficient algorithms are actively enriching existing video datasets with camera trajectories. This growing repository of pose-annotated data, like~\cite{zhou2025omniworld}, provides the reusable, large-scale foundation that our method can directly leverage.

Due to the inherent differences between egocentric and exocentric motion, we currently train a separate \ours~for each domain under our unified framework. A promising avenue for future work is to build a single, unified trajectory encoder. This could be achieved by introducing an explicit conditional domain token that allows the unified encoder to effectively distinguish and interpret the recorder's intent across both camera perspectives.

\CR{The utility of the camera trajectory signal intrinsically depends on the correlation between camera motion and semantic content. While this signal is highly informative for physical activities, it is naturally weaker for fine-grained procedural tasks characterized by subtle, localized motions. Consequently, \ours~can struggle to distinguish actions with kinematically similar patterns and inherently lacks the capacity to encode specific noun semantics. We present two failure modes of our investigation, as shown in Fig.~\ref{fig:failure_case}. The left panel reveals that \ours~struggles with subtle motion patterns (``press a button'' in this case), confuses it with the adjacent action of ``taking something from the countertop'', and inherently fails to encode specific noun semantics. The right panel highlights an issue with the pose source: we observe cases where the estimator (\eg, ViPE~\cite{huang2025vipe}) mistakenly correlates object motion with camera motion. This failure suggests that further algorithmic development in camera pose estimation is necessary to ensure robust semantic analysis. 

Lastly, our investigation is scoped to everyday human activity videos; settings where camera motion is decoupled from content (\eg, fixed cameras) fall outside scope.}


\end{document}